\documentclass[11pt,a4paper]{article} 
\usepackage[hyperref]{acl2021} 
\usepackage{times} 
\usepackage{latexsym} 
\usepackage{graphicx} 
\usepackage{float} 
\usepackage[inline]{enumitem} 
\usepackage{booktabs}

\usepackage{microtype}

\aclfinalcopy 
\def\aclpaperid{17} 

\def\figref#1{Figure~\ref{fig:#1}} 
\def\figlabel#1{\label{fig:#1}\label{p:#1}}

\def\tabref#1{Table~\ref{tab:#1}} \def\tablabel#1{\label{tab:#1}\label{p:#1}}
 \def\secref#1{\S\ref{sec:#1}}
\def\seclabel#1{\label{sec:#1}} \def\eqref#1{Eq.~\ref{eqn:#1}}

\def\compactsub#1{\textbf{#1.}}

\newcounter{notecounter} 
\newcommand{\enotesoff}{\long\gdef\enote##1##2{}}

\enotesoff
\long \def\eat#1{}

\def\parcour{ParCourE}
\def\multalign{\textsc{MultAlign}}
\def\lexicon{\textsc{Lexicon}}
\def\explore{\textsc{Interactive}}

 \def\simalign{SimAlign} 
\def\nlangs{1334} \def\neditions{1758} \def\nverses{20M}
\def\nversesexact{20,470,892} \def\avgverses{11,520} 
 \def\avgtokens{28.6}

\title{\parcour{}: A Parallel Corpus Explorer for\\
a Massively Multilingual Corpus}

\author{Ayyoob Imani$^1$, Masoud Jalili Sabet$^1$, Philipp Dufter$^1$, \\
	{\bf Michael Cysouw$^2$, Hinrich Sch\"{u}tze$^1$}\\
	$^1$Center for Information and Language Processing (CIS), LMU Munich, Germany\\
	$^2$Research Center Deutscher Sprachatlas, Philipps University Marburg, Germany.\\
	{\tt \{ayyoob, masoud, philipp\}@cis.lmu.de}}

\date{}

\begin{document} 
\maketitle 
\begin{abstract}
	With more than 7000 languages worldwide, multilingual natural language
	processing (NLP) is essential both from an academic and commercial
	perspective. Researching typological properties of languages is fundamental
	for progress in multilingual NLP. Examples include assessing language
	similarity for effective transfer learning, injecting inductive biases into
	machine learning models or creating resources such as dictionaries and
	inflection tables. We provide \parcour{}, an online tool that allows to browse
	a word-aligned parallel corpus, covering \nlangs{} languages. We give evidence
	that this is useful for typological research. \parcour{} can be set up for any
	parallel corpus and can thus be used for typological research on other corpora
	as well as for exploring their quality and properties. 
\end{abstract}

\section{Introduction}

While $\approx$7000 languages are spoken \cite{ethnologue}, the bulk
of NLP research addresses English only.
 However, multilinguality is an essential element of NLP. It not only supports exploiting common structures across languages
and eases maintenance for globally operating companies, but also helps save
languages from digital extinction and fosters more diversity
in NLP techniques.

There are extensive resources that can be used for massively multilingual
typological research, such as WALS \cite{wals}, Glottolog \cite{glottolog}, BabelNet
\cite{navigli2012babelnet} or http://panlex.org. Many
of them are manually created or crowdsourced, which guarantees high quality, but
limits  coverage, both in terms of content and languages. 

\begin{figure}[t] \centering 
	\includegraphics[width=0.8\linewidth]{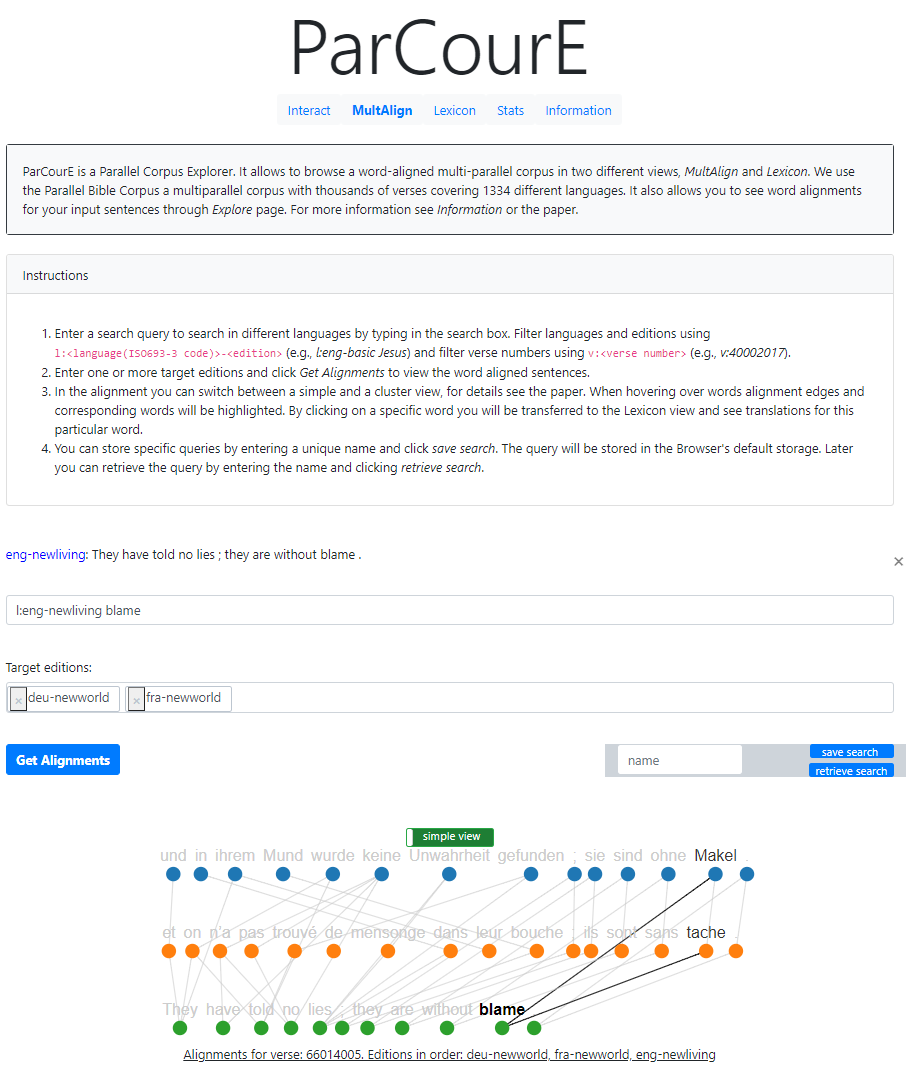} 
	\caption{Screenshot of the \parcour{} interface. It provides a word-aligned
	version of the Parallel Bible Corpus (PBC) spanning 1334 languages. Users can
	search for sentences in any language and see their alignments in other languages 
	from \multalign{} page.
	Alternatively they can feed their parallel sentences to \explore{} view 
	and see their word level alignments.
	They can look up translations of words in other languages, automatically induced
	from word alignments, from the \lexicon{} view (This page is interconnected with \multalign{}).
	 Statistics of the corpus is calculated and shown in the Stats view.}
	\figlabel{parcoure}
\end{figure}

We work on the Parallel Bible Corpus (PBC) \cite{mayer-cysouw-2014-creating},
 covering \nlangs{} languages. More specifically, we provide a
word-aligned version of PBC, created using state-of-the-art word alignment
tools. As word alignments themselves are only of limited use, we provide an
interactive online tool\footnote{\url{http://parcoure.cis.lmu.de/ }} that allows
effective browsing of the alignments.

The main contributions of this work are: 
\begin{enumerate*}
	[label={\textbf{\roman{*})}}] 
	\item We provide a word-aligned version of the Parallel Bible Corpus (PBC)
	spanning \nlangs{} languages and a total of \nverses{} sentences (`verses').
	For the alignment we use the state-of-the-art alignment methods SimAlign
	\cite{jalili-sabet-etal-2020-simalign} and Eflomal \cite{Ostling2016efmaral}. 
	\item We release \parcour{}, a user interface for
          browsing word alignments, see
the \multalign{} view in
          \figref{parcoure}. We
	demonstrate the usefulness of \parcour{} for typological research by
	presenting use cases in \secref{usecases}. 
	\item In addition to browsing word alignments, we provide an aggregated
	version in a \lexicon{} view and compute statistics that support assessing the
	quality of the word alignments. The two views (\multalign{} and \lexicon{}
	views) are interlinked, resulting in a richer user experience. 
	\item \parcour{} has a generic design and can be set up for any parallel
	corpus. This is useful for analyzing and managing parallel corpora; e.g.,
	errors in an automatically mined parallel corpus can be inspected and flagged
	for correction. 
\end{enumerate*}

\section{Related Work}

\textbf{Word Alignment} is an important tool for typological analysis
\cite{lewis-xia-2008-automatically} and annotation projection
\cite{yarowsky-etal-2001-inducing, ostling-2015-word,asgari-schutze-2017-past}. Statistical models such as IBM models
\cite{brown1993mathematics}, Giza++ \cite{och03:asc}, fast-align
\cite{dyer-etal-2013-simple} and Eflomal \cite{ostling2016efficient} are
widely used. Recently,  neural models were proposed, such as SimAlign
\cite{jalili-sabet-etal-2020-simalign}, Awesome-align \cite{dou2021word}, and methods that are based on neural machine translation \cite{garg-etal-2019-jointly,zenkel-etal-2020-end}. We use Eflomal and
SimAlign for generating alignments.

\textbf{Resources.} There are many online resources that enable typological
research. WALS \cite{wals} provides manually created features for
more than 2000 languages. We prepare a
multiparallel corpus for investigating these features on real data.
http://panlex.org is an online dictionary project with
2500 dictionaries covering 5700 languages and BabelNet
\cite{navigli2012babelnet} is a large semantic network covering 500 languages,
but their information is generally on the type level, without access to example
contexts. In contrast, \parcour{} supports the exploration
of  word
translations across \nlangs{} languages in context.

Another line of work
uses the
\textbf{Parallel Bible Corpus (PBC)}
for analysis.
\newcite{asgari-schutze-2017-past} investigate tense typology across PBC
languages. \newcite{xia-yarowsky-2017-deriving} 
created a multiway alignment based on fast-align
\cite{dyer-etal-2013-simple} and extracted resources such as
paraphrases for 27 Bible editions. \newcite{wu-etal-2018-creating} used alignments to extract names
from the PBC.

One of the first attempts to index the Bible and align words in multiple
languages were Strong's numbers \cite{strong2009strong}; they tag words with
similar meanings with the same ID. \citet{mayer-cysouw-2014-creating} created an
inverted index of word forms.
\citet{ostling2014bayesian} align 
massively parallel corpora simultaneously. 
We use the Eflomal word aligner by the same authors{ostling2016efficient}.

  Finally, we review work on \textbf{Word Alignment Browsers.}
\newcite{gilmanov-etal-2014-swift}'s tool supports
visualization and editing of word alignments.
\newcite{akbik-vollgraf-2017-projector}
use
co-occurrence weights for word alignment and
provide a tool for the
inspection of annotation projection. 
\newcite{aulamo-etal-2020-opusfilter}'s 
filtering tool increases the
quality of (mined) parallel corpora. 
\newcite{graen17multilingwis2} rely on linguistic
preprocessing, target corpus and word alignment
exploration, do not show the graph of alignment edges and do
not provide a dictionary view.
While there is commonality with this prior work,
\parcour{} is distinguished by both its functionality and
its motivating use cases:
an important use case for us are typological searches;
linguistic preprocessing is not
available for many PBC languages; 
\parcour{} can be used
as an interactive explorer (but is not
a fully-automated pipeline for a specific use case);
our goal is not annotation;
we use state-of-the-art word alignment methods.
However, much of the
complementary functionality in prior work would be useful additions to \parcour{}.
Another source of useful additional functionality would be
work on embedding learning
\cite{dufter-etal-2018-embedding,kurfali18pbc}
and machine translation
\citep{tiedemann2018emerging,santy-etal-2019-inmt,mueller-etal-2020-analysis}
for PBC.

\section{Features}

\parcour{}'s user facing functionality can be divided into
three main parts:
\multalign{} and \lexicon{} views and interconnections
between the two.

\begin{figure}
	[t] \centering 
	\includegraphics[width=1.0\linewidth]{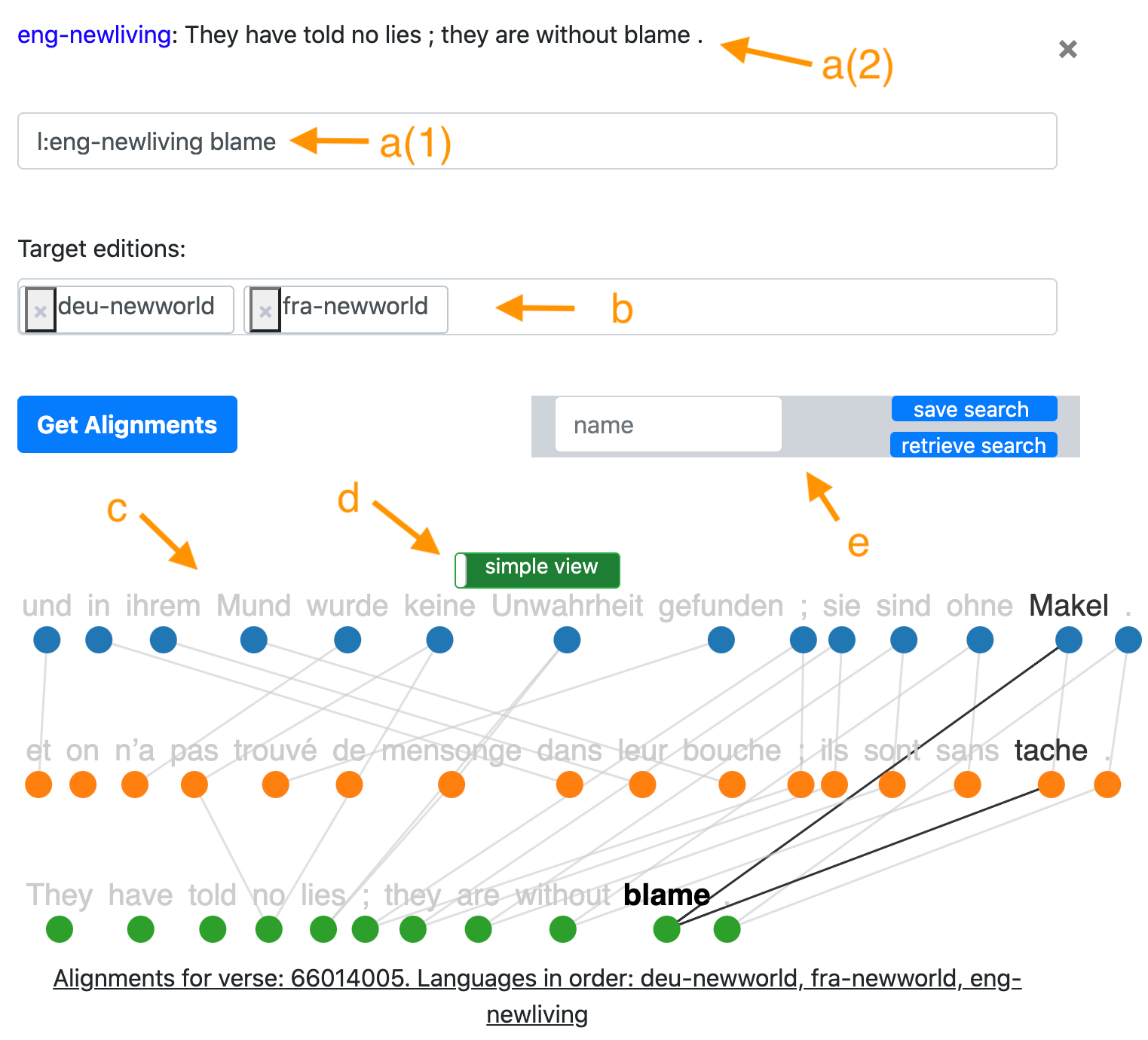} 
	\caption{An overview of the \multalign{} view. a) Search field for selecting
sentences [a(1)] and the list of selected sentences [a(2)]. Any language can be
used for the source sentence -- in this case, it is English. b) Search bar for
selecting the target languages. c) The alignment graph for the selected
sentences in the source and the target languages. d) Switch button for simple
view / cluster view. e) Save and retrieve search results}
\label{fig:multalign_overview} 
\end{figure}

\subsection{Multiparallel Alignment Browser: \multalign{}} 

\parcour{} allows the user to search through the parallel corpus and check word
alignments in a multiparallel corpus. An overview of \multalign{} is
shown in Figure \ref{fig:multalign_overview}.

In the \textbf{search field} (a(1)), the user can enter a text query and select (a(2)) multiple sentences for alignment. For
narrowing the search scope, the language and edition of the text segment can be
specified in the beginning, e.g., by typing
\emph{l:eng-newworld2013}. Similarly,
\emph{v:40002017} specifies a verse ID.

PBC has  \nlangs{}, so showing alignments for
all translations of a sentence is difficult. We provide a
drop-down (b) to select a subset of target languages
for display.

For each sentence, a graph of alignment edges between
selected languages is shown (c). By hovering over a word, the alignments of that word
will be highlighted. Above each alignment graph, there is a button to switch
between \textbf{Simple view} and \textbf{Cluster view} (d). In the simple view,
when hovering over a word, only the alignment edges connected to that word are
highlighted; in the cluster view, all words in a cluster (neighbors of neighbors) that are aligned
together will be highlighted.
We do not actually run any clustering algorithm on the
alignment graph. Instead we simply highlight words that are up to two
hops away from the hovered word. This
helps spot a group of words across languages that have the same
meaning.

Creating  queries for typology research can take
time. Thus, \multalign{} allows the user to \textbf{save and
  retrieve} (e) queries. \eat{The user saves the search
  scenario with a name and can later retrieve the results
  and continue exploring.}

\subsection{Lexicon View: \lexicon{}}
The \multalign{} view allows the user to focus on word alignments on the sentence
level and study the typological structure of languages in
context. The \lexicon{} view
focuses on word translations. The user can specify a source language by selecting the language code. This is
to distinguish words with the same spelling in different languages. 
The user can search for one or multiple word(s) and specify
target language(s).  A pie chart for each target
language depicting translations of the word is generated. \figref{lexicon_view} shows
German translations of ``confusion'' and
the number of alignment edges for each.
Word alignments are not perfect, so 
pie charts may also contain errors.

\begin{figure}[t] \centering 
	\includegraphics[width=0.75\linewidth]{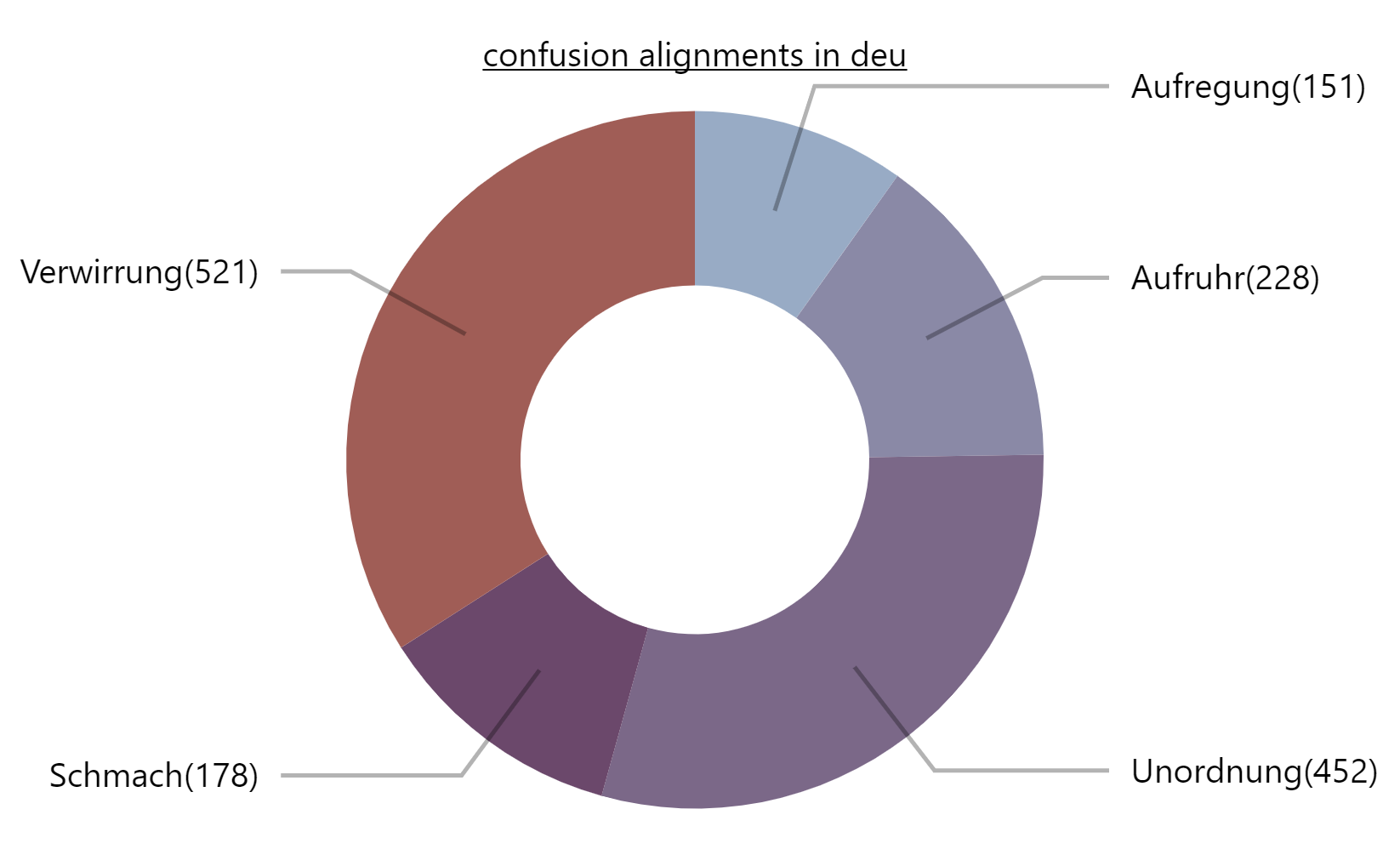} 
	\caption{\lexicon{} view example: for the English word ``confusion'', there
are five frequent translations in German. ``Unordnung'' literally means
``disorder'' and ``Verwirrung'' means ``bewilderment''.} \figlabel{lexicon_view}
\end{figure}

\subsection{Interconnections}

Both \multalign{} and \lexicon{} views provide
important features to the user for exploring the parallel corpus. For many use
cases (cf.\ \secref{usecases})\eat{ (e.g., error analysis of machine translation, see below)}, the user may
need to go back and forth between the views. For
example, if she notices an error in the word alignment, she may want to check
the \lexicon{} statistics to see if one of the typical translations of an
incorrectly aligned word occurs in the sentence.

\eat{To make going back and forth between \multalign{} and \lexicon{} views
efficient}Thus, the two views are interconnected. In the \multalign{} view, the user
will be transferred to the \lexicon{} statistics of a word by clicking on it. This will open the
\lexicon{} view, showing the search results for the selected word. Conversely,
 if the user clicks on one of the target translations in the \lexicon{} view,
the \multalign{} view will show sentences where this correspondence
is part of the word alignment between source and target translation.

\subsection{Alignment Generation View: \explore{}}
The views mentioned so far provide the ability to search
over the indexed corpus.  This is useful when
the main corpus of interest is fixed and the user has
generated its alignments.

The \explore{} view allows the user to study the alignments between arbitrary input sentences that are not necessarily in the corpus.
Since the input sentences are not part of a corpus, \explore{} uses SimAlign to generate alignments for all possible pairs of sentences.
Similar to \multalign{}, the \explore{} view shows the
alignment between the input sentences.

\section{Experimental Setup}

\compactsub{Corpus}
We set up \parcour{} on the PBC corpus provided by
\citet{mayer-cysouw-2014-creating}. The version we use consists of \neditions{}
editions (i.e., translations) of the Bible in \nlangs{} 
languages (distinct ISO 639-3 codes).
\tabref{pbc_statics}
shows corpus statistics.
We use the PBC tokenization,
which contains errors for a few languages (e.g., Thai). We
extract word alignments for all possible language pairs. Since not all Bible
verses are available in all languages, for each language pair we only consider
mutually available verses. 

PBC aligns Bible editions  on the verse
  level by using verse-IDs that indicate
book, chapter and verse (see below). Although one verse may contain multiple sentences, we do not split
verses into individual sentences and consider each verse as one sentence. \eat{We use
verse-ID as sentence number for \parcour{}.}

\begin{table}
	\centering \footnotesize 
	\begin{tabular}{lr|lr}
		\midrule
		 \# editions & \neditions{} &		\# verses & \nversesexact{} \\

		\# languages & \nlangs{} &		\# verses / \# editions & \avgverses{} \\

                &&		\# tokens / \# verses & \avgtokens{} \\
		\midrule
	\end{tabular}
	\caption{PBC corpus statistics\tablabel{stats}} \tablabel{pbc_statics}
\end{table}

\compactsub{Retrieval}
Elasticsearch\footnote{\url{https://www.elastic.co/}} is a
fast and scalable open source search engine that provides
distributed fulltext search.  The setup is straightforward
using an easy-to-use JSON web interface. We use it as the
back-end for \parcour's search requirement. We find that a
single instance is capable of handling the whole PBC corpus
efficiently, so we do not need a distributed setup. For
bigger corpora, a distributed setup may be required. We
created two types of inverted indices for our data: an
edge-ngram index to support search-as-you-type capability
and a standard index for normal queries.

\compactsub{Alignment Generation} 
\simalign{} \cite{jalili-sabet-etal-2020-simalign} is a recent word alignment
method that uses representations from pretrained language models to align sentences. 
It has achieved better results than statistical word aligners. 
For the languages that multilingual BERT \cite{devlin-etal-2019-bert} supports, we use
\simalign{} to generate word alignments. For the remaining languages, we use
Eflomal \cite{Ostling2016efmaral}, an efficient word aligner using a Bayesian
model with Markov Chain Monte Carlo (MCMC) inference. The alignments generated
by \simalign{} are symmetric. We use 
atools\footnote{\url{https://github.com/clab/fast_align}}
and the grow-diag-final-and heuristic
to symmetrize
Eflomal alignments.

\compactsub{Lexicon Induction} 
We exploit the generated word alignments to
induce lexicons for all 889,111 \eat{$(= 1334(1334-1)/2)$ possible} language pairs. To
this end, we consider aligned words as translations of each other. For a given
word from the source language, we count the number of times a word from the
target language is aligned with it. The higher the number of alignments between
two words, the higher the probability that the two have the same meaning. We 
filter out translations with frequency less than 5\%.

\section{Backend Design} 

\eat{In this section, we describe the tools that prepare the
input data for \parcour{}. }An overview of our architecture can be found in \figref{system}. The code is available online.\footnote{\url{https://github.com/cisnlp/parcoure}}

\begin{figure}
	\centering 
	\includegraphics[width=1.0\linewidth]{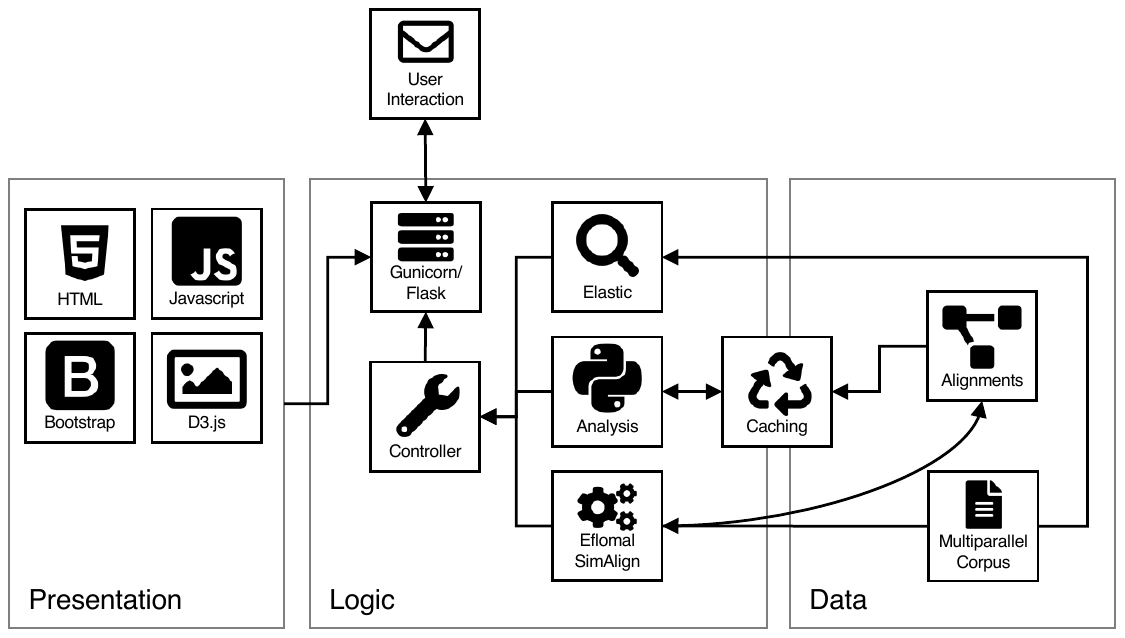} 
	\caption{Overview of the system architecture. We use a standard front-end
	stack with d3.js for visualization. The backend is written in Python, which we
	use for computing alignments and performing analyses such as lexicon
	induction. We use Elasticsearch for search. The input is
	a multiparallel corpus for which all alignments are precomputed. For speeding
	up the system we use smart caching algorithms for our analyses. Icons taken
	without changes from \url{https://fontawesome.com/license}.\figlabel{system}} 
\end{figure}

\compactsub{Parallel Data Format} 
We use the PBC corpus format 
\citep{mayer-cysouw-2014-creating}: each verse has a unique ID 
across languages / editions, the \emph{verse-ID}. The verse-ID is an 8-digit number, consisting of
two digits for the book (e.g., 41 for the Gospel of Mark), three digits
for the Chapter, and two digits for the verse itself. There are separate files
for each edition. In each edition file, a line consists of the ID and the verse,
separated by a tab.

\compactsub{Indexing}
We identify 
\eat{In a parallel corpus, sentences (`verses') with the same identifier are
translations of each other. For some languages the PBC includes multiple
translations per language. We call each of these translations an edition. We
identify }a PBC verse using the following format:  $\{$verse-ID$\}$@$\{$language-code$\}$-$\{$edition-name$\}$. We use this
identifier to save and retrieve sentences with Elasticsearch. In addition, we store all metadata
identifiers within Elasticsearch. Thus, we can search for a sentence by keyword,
sentence number (= verse-ID), language code, or edition name. 

\parcour{} also supports the Corpus Alignment Encoding (CES)\footnote{\url{
https://www.cs.vassar.edu/CES/}} format. One can download parallel
corpora in CES format and use our tools to adapt them to \parcour's input
format. \eat{The user can then set up \parcour{} for the resulting data.}

\compactsub{Alignment Computation}
Since Eflomal's performance depends on the amount of data it uses for training,
we concatenate
all editions to create a bigger training corpus for languages that have more than one edition. \eat{We also align all available
editions of the same sentence from the two languages. }If language
$l_1$ has two, and language $l_2$ three
different editions, then the final training corpus for this language pair will
contain six aligned edition pairs.

\compactsub{System Architecture}
\parcour{} is built on top of modern open source technologies, see \figref{system}. The
back-end uses the Flask web
framework,\footnote{\url{https://flask.palletsprojects.com}} Gunicorn web
server,\footnote{\url{https://gunicorn.org/}} and Elasticsearch.\footnote{\url{https://www.elastic.co/}} The front\-end utilizes the
Bootstrap CSS framework,\footnote{\url{https://getbootstrap.com/}} and the d3
visualization library.\footnote{\url{https://d3js.org/}} Since all these tools
are free and open-source, there is no restriction on setting up and releasing a
new \parcour{} instance. To extract word alignments, one can use 
any tool, such as Eflomal, fast\_align or SimAlign.
\eat{Eflomal,\eat{\footnote{\url{https://github.com/robertostling/Eflomal}}}
fast\_align,\footnote{\url{https://github.com/clab/fast_align}} and
SimAlign\footnote{\url{https://github.com/cisnlp/simalign}}}

\compactsub{Performance Improvements} 
For good run-time performance, we precompute the
word alignments. Regarding \lexicon{}, given a query word and a target language, \parcour{}
first looks for a precomputed lexicon file; if it does not exist, \parcour{} 
obtains the translations for the query word online. To accelerate the translation
process, \parcour{} employs Python's multiprocessing library. The
number of CPU cores is decided online based on the number of editions
available for source and target languages.

For a corpus with \nlangs{}  languages, we will end up with 890,445 \eat{$(=
1334(1334-1)/2 + 1334)$ }alignment files and the same number of
lexicon files.
We cache alignment / lexicon files to speed up access. We use the 
Last Recently Used (LRU) cache replacement algorithm.

\section{\parcour{} Use Cases} \seclabel{usecases} 

Languages differ in how they encode meanings/functions. There are various
aspects that make such differences an interesting problem when dealing with a
dataset that has good coverage of the entire variation of the world's languages.
(i) Many such differences between languages are not widely acknowledged in
linguistic theory, so to document the extent of variation becomes a discovery of
sorts. For example, the fact that interrogative words might distinguish between
singular and plural (\figref{gramdiff}) turns out to be a typologically salient
differentiation \citep{mayer2012}. (ii) The variation of linguistic marking is
even stronger in the domain of grammatical function, like the differentiation
between the interrogative and relative pronoun in \figref{gramdiff}. (iii) In
lexical semantics, \parcour{} supports the investigation of how languages carve
up the meaning space differently (cf.\ \figref{lexdiff}), especially when it
comes to the $\approx$1000 low-resource languages covered in PBC. Massively
parallel texts are an ideal resource to investigate such variation
\citep{haspelmath2003}.

\begin{figure}[t]
	\centering
	\includegraphics[width=0.95\linewidth]{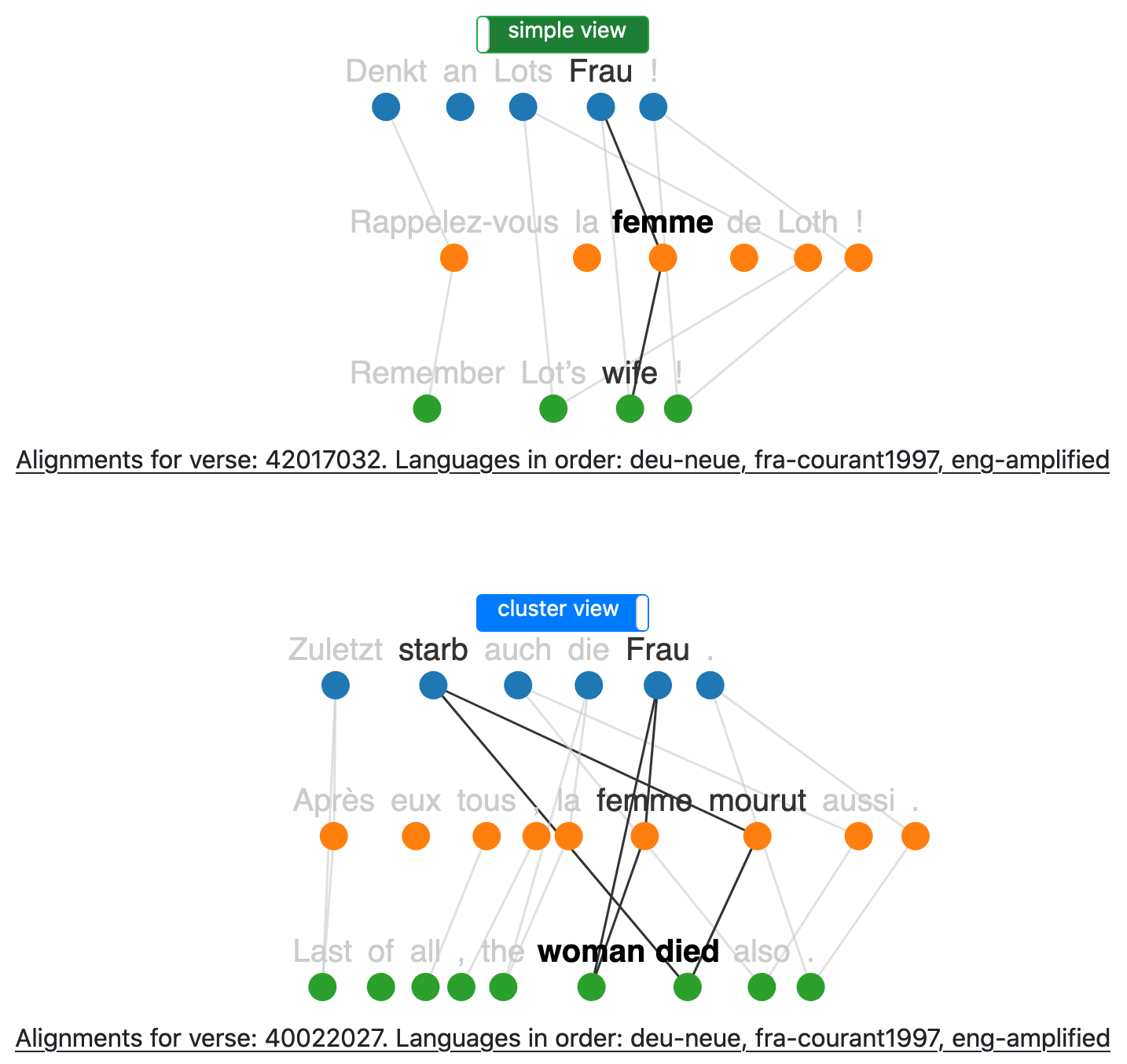}
	\caption{Use case 1,
		\emph{lexical differentiation}. French
		``femme'' has two different translations in English
		(``wife'' and ``woman'') whereas German also conflates the
		two different meanings.
	\figlabel{lexdiff}}
\end{figure}

\begin{figure}[t]
	\centering 
	\includegraphics[width=1.0\linewidth]{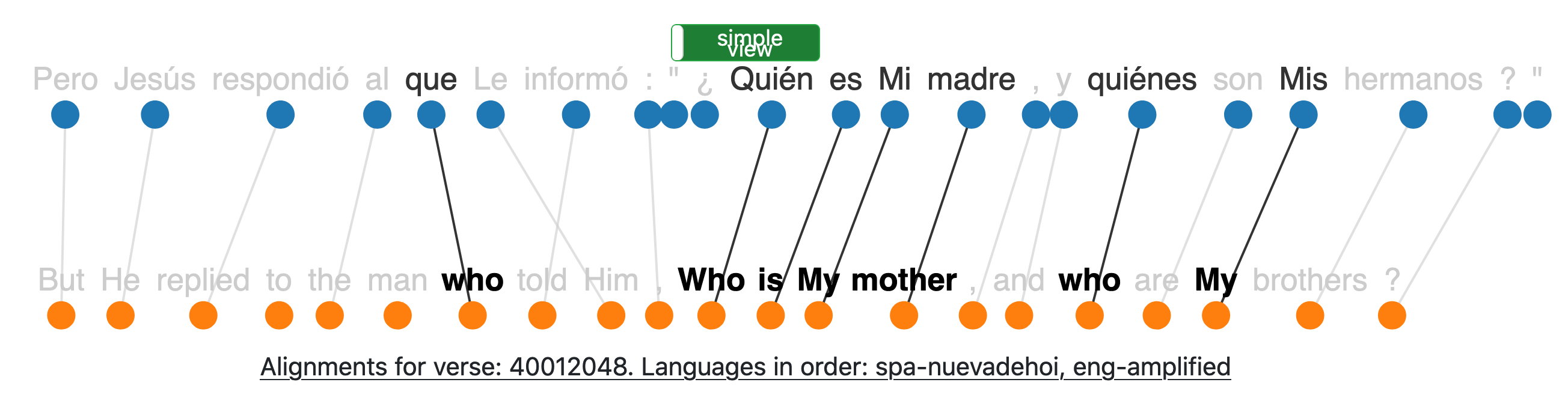} 
	\caption{Use case 2, 
		\emph{grammatical differentiation}. English ``who'' has three different
		translations in  this Spanish example:  relative pronoun (``que''), and 
		singular (``qui\'{e}n)'' and plural (``qui\'{e}nes'') interrogative
		pronoun.
	\figlabel{gramdiff}} 
\end{figure}

Grammatical differences between languages, like differences in word order, have
a long history in research on worldwide linguistic variation
\citep{greenberg1966, dryer1992}. However, being able to look at the usage of
word order in specific contexts (and being able to directly compare exactly the
same context across languages) is only possible by using parallel texts. For
example, specific orders of more than two elements can be directly extracted
from the parallel texts, like the order of demonstrative, numeral and noun
``these two commandments'' in \figref{ordervar} \citep{cysouw2010}.

\begin{figure}[t]
	\centering 
	\includegraphics[width=1.0\linewidth]{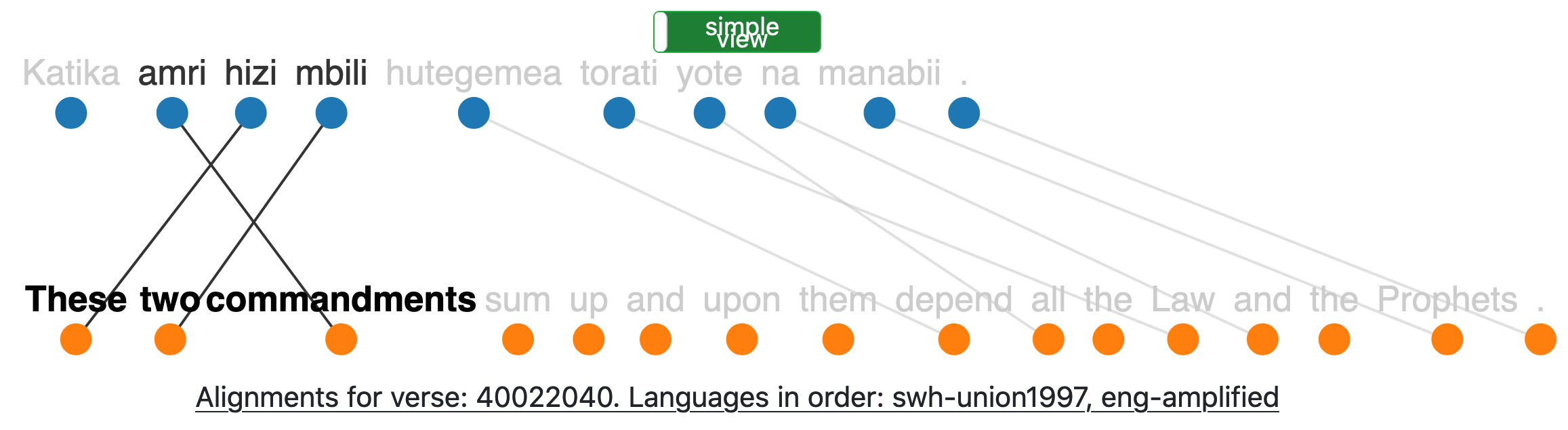} 
	\caption{Use case 3,
		\emph{word order variation}. The English order is demonstrative, numeral, 
		noun whereas Swahili has noun, demonstrative, numeral.
	\figlabel{ordervar}} 
\end{figure}

For lack of space, we describe four more use cases only briefly: grammatical
markers vs.\ morphology as devices to express grammatical features
(\figref{crsti}); differences in how languages use grammatical case
(\figref{usecasemorph}, ablative/dative in Latin can correspond to five
different cases in Croatian); and exploration of paraphrases
(\figref{paraphrase}). See the captions of the figures for more details. 

\begin{figure}[t]
	\centering
	\includegraphics[width=0.75\linewidth]{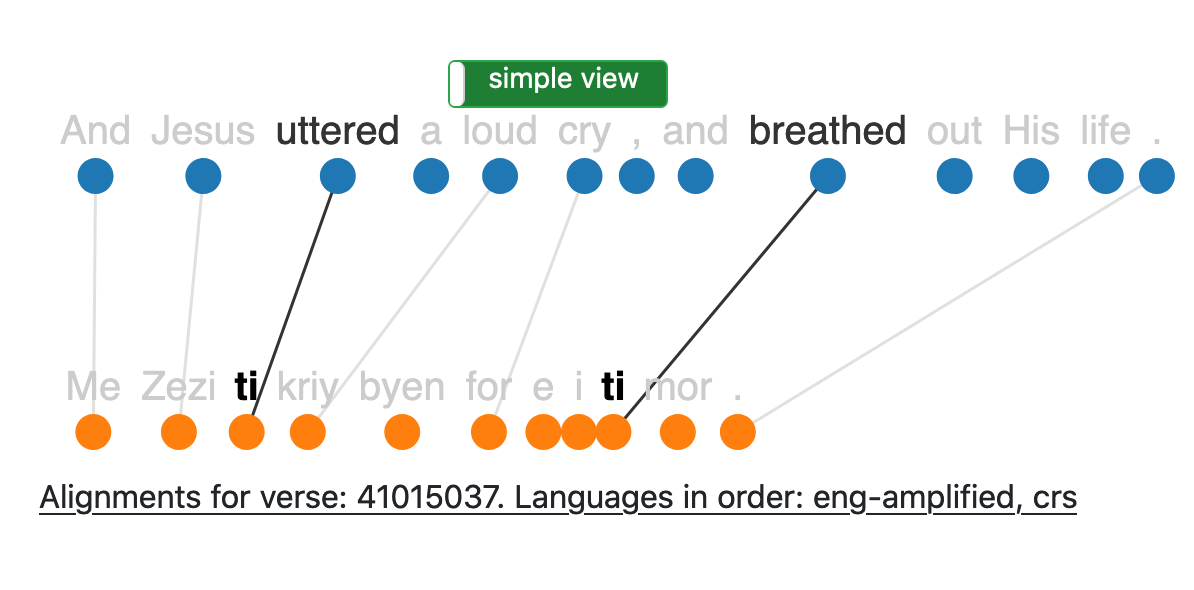}
	\caption{Use case 4,
    \emph{grammatical markers}.
		In contrast to English, Seychelles Creole does not inflect
		verbs for tense and uses the past
		tense marker ``ti'' instead.
	\figlabel{crsti}}
\end{figure}

\begin{figure}[t]
	\centering
	\includegraphics[width=0.75\linewidth]{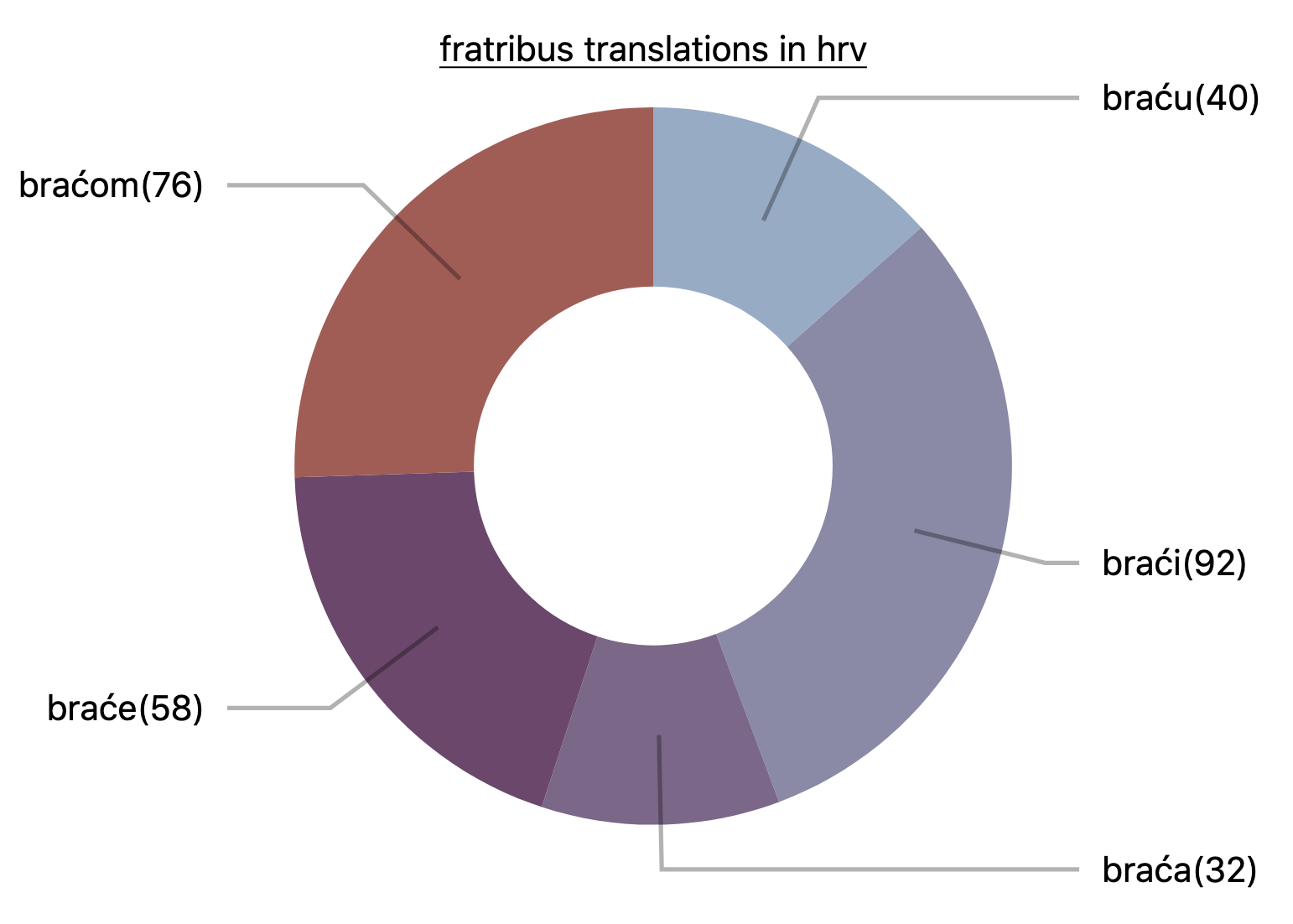}
	\caption{Use case 5, 
		\emph{morphology}. The Latin ending ``ibus'' in
    ``fratribus'' (dative/ablativ plural) corresponds to five
    different cases in Croatian: accusative, locative/dative,
    nominative, genitive, instrumental (clockwise starting from
    ``bra\'{c}u'').
	\figlabel{usecasemorph}}
\end{figure}

\begin{figure}[h]
	\centering 
	\includegraphics[width=1.0\linewidth]{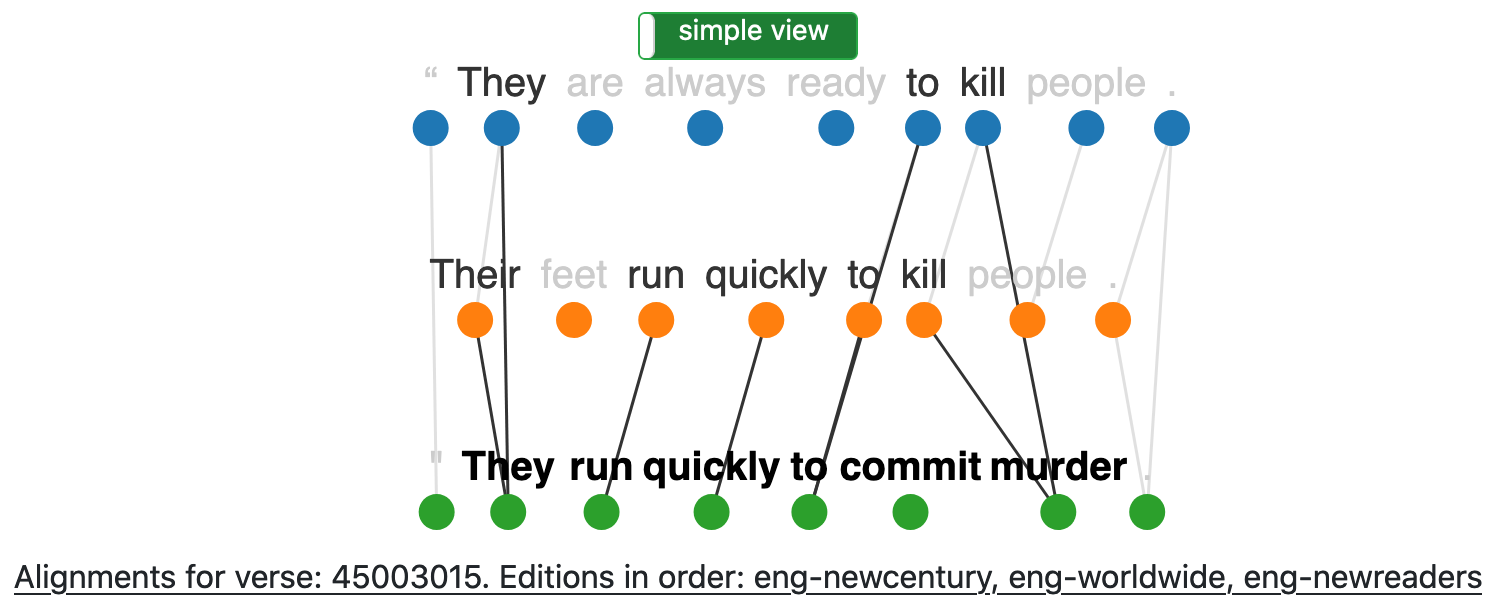} 
	\caption{Use case 6,
		\emph{paraphrases}. PBC is a rich source of paraphrases
		since high-resource languages have several translations (32 for English).
		\parcour{} can be used to explore these paraphrases. Here, the paraphrases
		``kill'' and ``murder'' are correctly aligned, ``always ready'' and ``run
		quickly'' are not.
	\figlabel{paraphrase}}
\end{figure}

\begin{figure}
	\centering 
	\includegraphics[width=1.0\linewidth]{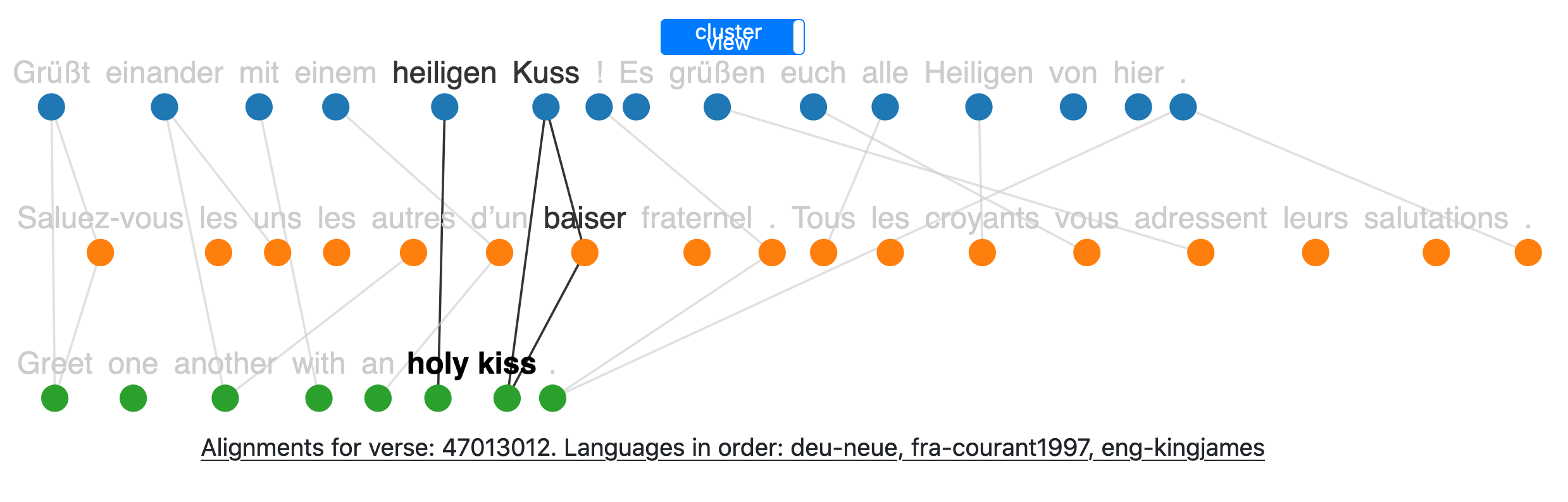} 
	\caption{Use case 7, \emph{quality analysis}. \parcour{} makes it easy to
	analyze the quality of the parallel corpus. For this sentence, part of a Bible
	verse present in German and French is missing in English. Note that the
	alignment of \textit{holy, heiligen} to French \textit{fraternel} is not
	discovered.
	\figlabel{holykiss}} 
\end{figure}

\section{Extension to Other Corpora} 

Our code is available on GitHub and can be
generically applied: you can create a \parcour{} instance for your own parallel
corpus. Parallel corpora are essential for machine translation (MT);
\parcour{}'s functionality is useful for analyzing the quality of a parallel
corpus and the difficulty of the translation problem it
poses. We give three examples
\begin{enumerate*}[label={\textbf{\roman{*})}}]
	\item Incorrect sentence alignments can be identified, e.g., cases in which a
	target sentence is matched with the merger of two sentences in the source:
	cf.\ \figref{holykiss} where a short sentence in English is aligned with
	German and French sentences that also contain a second sentence that is
	missing in English. This functionality is particularly helpful for
	\eat{automatically} mined parallel corpora that tend to contain erroneous
	sentence pairs. 
      \item Suppose an MT system trained on the parallel corpus makes a lexical
	error in a particular context $c$ by mistranslating source word $w_s$ 
	with target word $w_t$. The \lexicon{} view can be consulted for $w_s$ and the
	user can then click on the erroneous target word $w_t$ to get back to a
	\multalign{} view of aligned sentence pairs containing $w_s$ and $w_t$. 
	She can then analyze why the MT system  mismatched $c$ with these
	contexts. Examples of the desired translation are easy to find and inspect to
	support the formation of hypotheses as to the source
        of the error.
      \item For multi-source approaches to MT \cite{zoph-knight-2016-multi,
	firat-etal-2016-zero, libovicky2017attention, crego-etal-2010-local},
	\parcour{} supports the inspection of all input sentences together. The MT
	system output can also be loaded into \parcour{} for a view that contains all
	input sentences and the output sentence. Since any of the input sentences can
	be responsible for an error in multi-source MT, this facilitates analysis and
	hypothesis formation as to what caused a specific error.
	\end{enumerate*}

\subsection{Computing Infrastructure and Runtime}
We did all computations on a machine with 48 cores of Intel(R) Xeon(R) CPU E7-8857 v2 with 1TB memory.
In this experiment only one core was used.

We created a corpus of 5 translations in 4 languages, with around 31k parallel sentences (overally 155k sentences) 
and applied the \parcour{} pipeline to it.
Runtimes for different parts of the pipeline are reported in \tabref{runtimes}.
The installation of the package is straightforward and as shown in the table, it takes around 12 minutes to initiate \parcour{} on a small corpus with 4 languages.

\begin{table}[H]
	\footnotesize
	\centering
	\begin{tabular}{l|r}
		Method & Runtime \\
		\midrule
		Conversion from CES to \parcour{} format & 153 \\
		Indexing with Elasticsearch & 14 \\
		Alignment generation with Eflomal & 537  \\
		Stats calculation &  22 \\ 
		\midrule
		Overall & 726 \\
	\end{tabular}
	\caption{Runtime in seconds for each part of the pipeline to initiate a \parcour{} instance on a corpus with 4 languages and 31K parallel sentences. \tablabel{runtimes}}
\end{table}

\section{Conclusion}

Progress in multilingual NLP is an important goal of NLP
and requires researching typological properties of languages. Examples
include assessing language similarity for effective transfer learning, injecting
inductive biases into machine learning models and creating resources such as
dictionaries and inflection tables. To serve such use cases, we have created
\parcour{}, an online tool for browsing a word-aligned parallel corpus of
\nlangs{} languages, and given evidence that it is useful for typological
research. \parcour{} can be set up for any other parallel corpus, e.g.,
for quality control and improvement of automatically mined parallel corpora.

\section*{Acknowledgments} 

This work was supported by the European Research Council
(ERC, Grant No.\ 740516) and the German Federal Ministry of Education and
Research (BMBF, Grant No.\ 01IS18036A). The third author
was supported by the Bavarian research institute for digital
transformation (bidt) through their fellowship program. We
thank the anonymous reviewers for their constructive
comments.

\section{Ethical Considerations}
Word
alignments and lexicon induction as tasks themselves may not
have ethical implications.
However, working on a biblical corpus requires special consideration of the
following issues.

\begin{enumerate*}[label={\textbf{\roman{*})}}]
	\item The Bible is the central religious text of Christianity and the Hebrew
  Bible that of Judaism. It contains strong opinions and  world views
  (e.g., on divorce and homosexuality) that are not generally shared. We
  would like to emphasize that we treat the PBC simply as a
  multiparallel corpus, and the corpus does not necessarily reflect the
  opinions of the authors nor of the institutions funding the authors. 
	\item In a similar vein, while the PBC has  great language coverage and allows
	for typological analysis, we need to be aware that languages might not be
	accurately and completely reflected in the PBC. The language used in the PBC might be
	outdated and is restricted to a relatively small subset of topics and
	thus cannot be considered a balanced and complete view of the language.
	\item We also need to be aware of selection bias. The PBC only covers a
  subset of the world's languages. The selection criteria are unknown
  and may be based on historical and cultural biases that we are not
  able to assess.
\end{enumerate*}

\bibliographystyle{acl_natbib} 
\bibliography{anthology,acl2021}

\end{document}